\documentclass[11pt,a4paper]{article}
\usepackage[hyperref]{acl2019}
\usepackage{times}
\usepackage{latexsym}
\usepackage{danudefs}
\usepackage{algorithmic}
\usepackage{algorithm}
\usepackage{color}
\usepackage{booktabs}
\usepackage{xspace}
\usepackage{url}
\usepackage{mathtools}
\usepackage{amsmath}
\usepackage{graphicx}
\usepackage{subcaption}
\usepackage{multirow}
\usepackage{enumerate}
\usepackage{esvect}

\aclfinalcopy % Uncomment this line for the final submission

\title{Gender-preserving Debiasing  for Pre-trained Word Embeddings}

\author{
    Masahiro Kaneko\\
    Tokyo Metropolitan University, Japan\\
  {\tt kaneko-masahiro@ed.tmu.ac.jp}
    \And
    Danushka Bollegala \\
  University of Liverpool, UK\\
  {\tt danushka@liverpool.ac.uk}}

\date{}

\begin{document}
\maketitle

\begin{abstract}
Word embeddings learnt from massive text collections have demonstrated significant levels of discriminative biases such as gender, racial or ethnic biases, which in turn bias the down-stream NLP applications that use those word embeddings.
Taking gender-bias as a working example, we propose a debiasing method that preserves non-discriminative gender-related information, while removing stereotypical discriminative gender biases  from pre-trained word embeddings.
Specifically, we consider four types of information: \emph{feminine}, \emph{masculine}, \emph{gender-neutral} and \emph{stereotypical}, which represent the relationship between gender vs. bias, and propose a debiasing method that
(a) preserves the gender-related information in feminine and masculine words,
(b) preserves the neutrality in gender-neutral words, and
(c) removes the biases from stereotypical words.
Experimental results on several previously proposed benchmark datasets show that our proposed method can debias pre-trained word embeddings better than existing SoTA methods proposed for debiasing word embeddings while preserving gender-related but non-discriminative information.
\end{abstract}

\section{Introduction}
\label{sec:intro}

% Explain word embeddings are impressive but have biases. They make downstream NLP systems biased.
% legal, business and ethical arguments for debiasing
% limitations of prior work
% novelty and advantages of the proposed method
% describe the proposed method in brief and its performance

Despite the impressive success stories behind word representation learning~\cite{BERT,Elmo,Glove,Mikolov:NAACL:2013,Milkov:2013}, further investigations into the learnt representations have revealed several worrying issues. The semantic representations learnt, in particular from social media, have shown to encode significant levels of racist, offensive and discriminative language usage~\cite{Tolga:NIPS:2016, Zhao:2018ab,Elazar:EMNLP:2018,Rudinger:2018aa,Zhao:2018aa}.
For example, \newcite{Tolga:NIPS:2016} showed that word representations learnt from a large (300GB) news corpus to amplify unfair gender biases. Microsoft's AI chat bot \emph{Tay} learnt abusive language from Twitter within the first 24 hours of its release, which forced Microsoft to shutdown the bot~\cite{Tay}.  
\newcite{WEAT} conducted an implicit association test (IAT)~\cite{IAT} using the cosine similarity measured from word representations, and showed that word representations computed from a large Web crawl contain human-like biases with respect to gender, profession and ethnicity.

Given the broad applications of pre-trained word embeddings in various down-stream NLP tasks such as 
machine translation~\cite{Zou:EMNLP:2013}, sentiment analysis~\cite{Shi:2018aa}, dialogue generation~\cite{Zhang:2018ab} etc., it is important to debias word embeddings \emph{before} they are applied in NLP systems that interact with and/or make decisions that affect humans.
We believe that no human should be discriminated on the basis of demographic attributes by an NLP system, and 
there exist clear legal~\cite{TreatyofAmsterdam}, business and ethical obligations to make NLP systems unbiased~\cite{Holstein:2018}.

Despite the growing need for unbiased word embeddings, debiasing pre-trained word embeddings is a challenging task that requires a fine balance between removing information related to discriminative biases, while retaining  information that is necessary for the target NLP task.
For example, profession-related nouns such as \emph{professor}, \emph{doctor}, \emph{programmer} have shown to be stereotypically male-biased, whereas \emph{nurse}, \emph{homemaker} to be stereotypically female-biased, and a debiasing method must remove such biases. 
On the other hand, one would expect\footnote{This indeed is the case for pre-trained GloVe embeddings}, \emph{beard} to be associated with male nouns and \emph{bikini} to be associated with female nouns, and preserving such gender biases would be useful, for example, for a recommendation system~\cite{garimella-banea-mihalcea:2017:EMNLP2017}. 
As detailed later in \autoref{sec:related}, existing debiasing methods can be seen as embedding word embeddings into a subspace that is approximately orthogonal to a gender subspace spanned by gender-specific word embeddings.
Although unsupervised, weakly-supervised and adversarially trained models have been used for learning such embeddings, they primarily focus on the male-female gender direction and ignore the effect of words that have a gender orientation but not necessarily unfairly biased. 

To perform an extensive treatment of the gender debiasing problem, we split a given vocabulary $\cV$ into four mutually exclusive sets of word categories: 
(a) words $w_{f} \in \cV_{f}$ that are female-biased but non-discriminative,
(b) words $w_{m} \in \cV_{m}$ that are male-biased but non-discriminative,
(c) words $w_{n} \in \cV_{n}$ that are gender-neutral, and 
(d) words $w_{s} \in \cV_{s}$ that are stereotypical (i.e., unfairly\footnote{We use the term \emph{unfair} as used in \emph{fairness-aware machine learning.}} gender-biased).
Given a large set of pre-trained word embeddings and small seed example sets for each of those four categories, we learn an
embedding that
(i) preserves the feminine information for the words in $\cV_{f}$, 
(ii) preserves the masculine information for the words in $\cV_{m}$, 
(iii) protects the neutrality of the gender-neutral words in $\cV_{n}$, while
(iv) removing the gender-related biases from stereotypical words in $\cV_{s}$.
The embedding is learnt using an encoder in a denoising autoencoder, while the decoder is trained to reconstruct the original word embeddings from the debiased embeddings that do not contain unfair gender biases.
The overall model is trained end-to-end to dynamically balance the competing criteria (i)-(iv).

We evaluate the bias and accuracy of the word embeddings debiased by the proposed method on multiple benchmark datasets.
On the \textsf{SemBias}~\cite{Zhao:2018ab} gender relational analogy dataset, our proposed method outperforms previously proposed \emph{hard-debiasing}~\cite{Tolga:NIPS:2016} and \emph{gender-neural Global Vectors} (GN-GloVe)~\cite{Zhao:2018ab} by correctly debiasing stereotypical analogies.
Following prior work, we evaluate the loss of information due to debiasing on benchmark datasets for semantic similarity and word analogy. 
Experimental results show that the proposed method can preserve the semantics of the original word embeddings, while removing gender biases. This shows that the debiased word embeddings can be used as drop-in replacements for word embeddings used in NLP applications.
Moreover,  experimental results show that our proposed method can also debias word embeddings that are already debiased using previously proposed debiasing methods such as GN-GloVe to filter out any remaining gender biases, while preserving semantic information useful for downstream NLP applications. This enables us to use the proposed method in conjunction with existing debiasing methods.

\section{Related Work}
\label{sec:related}

To reduce the gender stereotypes embedded inside pre-trained word representations, \newcite{Tolga:NIPS:2016}~proposed a post-processing approach that projects gender-neutral words to a subspace, which is orthogonal to the gender dimension defined by a list of gender-definitional words.
They refer to words associated with gender (e.g., \emph{she}, \emph{actor}) as gender-definitional words, and the remainder gender-neutral.
They proposed a \emph{hard-debiasing} method where the gender direction is computed as the vector difference between the embeddings of the corresponding gender-definitional words, and a \emph{soft-debiasing} method, which balances the objective of preserving the inner-products between the original word embeddings, while projecting the word embeddings into a subspace orthogonal to the gender definitional words. 
They use a seed set of gender-definitional words to train a support vector machine classifier, and use it to expand the initial set of gender-definitional words.
Both hard and soft debiasing methods ignore gender-definitional words during the subsequent debiasing process, and focus only on words that are \emph{not} predicted as gender-definitional by the classifier. 
Therefore, if the classifier erroneously predicts a stereotypical word as a gender-definitional word, it would not get debiased.

\newcite{Zhao:2018ab}~proposed Gender-Neutral Global Vectors (GN-GloVe) by adding a constraint to the Global Vectors (GloVe)~\cite{Glove} objective such that the gender-related information is confined to a sub-vector. During optimisation, the squared $\ell_{2}$ distance between gender-related sub-vectors are maximised, while simultaneously minimising the GloVe objective.
GN-GloVe learns gender-debiased word embeddings from scratch from a given corpus, and cannot be used to debias pre-trained word embeddings.
Moreover, similar to hard and soft debiasing methods described above, GN-GloVe uses pre-defined lists of feminine, masculine and gender-neutral words and \emph{does not} debias words in these lists.

Debiasing can be seen as a problem of \emph{hiding} information related to a \emph{protected} attribute such as gender, for which adversarial learning methods~\cite{Xie:NIPS:2017,Elazar:EMNLP:2018,Li:2018ab} have been proposed in the fairness-aware machine learning community~\cite{Faisal:2009}. In these approaches, inputs are first encoded, and then two classifiers are trained -- a \emph{target task predictor} that uses the encoded input to predict the target NLP task, and a \emph{protected-attribute predictor} that uses the encoded input to predict the protected attribute. The two classifiers and the encoder is learnt jointly such that the accuracy of the target task predictor is maximised, while minimising the accuracy of the protected-attribute predictor. 
However, \newcite{Elazar:EMNLP:2018} showed that although it is possible to obtain chance-level development-set accuracy for the protected attribute during training, a post-hoc classifier, trained on the encoded inputs can still manage to reach substantially high accuracies for the protected attributes. They conclude that adversarial learning alone does not guarantee invariant representations for the protected attributes.

Gender biases have been identified in several tasks in NLP such as coreference~\cite{Rudinger:2018aa,Zhao:2018aa} resolution and machine translation~\cite{Prates:2018}.
For example, rule-based, feature-based as well as neural coreference resolution methods trained on biased resources have shown to reflect those biases in their predictions~\cite{Rudinger:2018aa}.
Google Machine Translation, for example, provides male and female versions of the translations\footnote{\url{https://bit.ly/2B0nVHZ}},
when the gender in the source language is ambiguous.

\section{Gender-Preserving Debiasing}
\label{sec:method}

% 4 types, revisit
% objective function
% optimisation

% TODO
% Tell that we do not need gender paired words for the first two objectives but only for the third one.
% Tell what would happen if words are not mutually exclusive (stereotypes take precedence)
% Tell how the method can be extended to other protected attributes (ethnicity for example)

\subsection{Formulation}
Given a pre-trained set of $d$-dimensional word embeddings $\{\vec{w}_{i}\}_{i=1}^{|\cV|}$, over a vocabulary $\cV$, we consider the problem of learning a map $E: \R^{d} \rightarrow \R^{l}$ that projects the original pre-trained word embeddings to a debiased $l$-dimensional space.
We do not assume any knowledge about the word embedding learning algorithm that was used to produce the pre-trained word embeddings given to us. Moreover, we do not assume the availability or access to the language resources such as corpora or lexicons that might have been used by the word embedding learning algorithm.
Decoupling the debiasing method from the word embedding learning algorithm and resources increases the applicability of the proposed method, enabling us to debias pre-trained word embeddings produced using different word embedding learning algorithms and using different types of resources.

We propose a debiasing method that models the interaction between the values of the protected attribute (in the case of \emph{gender} we consider \emph{male}, \emph{female} and \emph{neutral} as possible attribute values), and whether there is a stereotypical bias or not. 
Given four sets of words: \emph{masculine} ($\cV_{m}$), \emph{feminine} ($\cV_{f}$), \emph{neutral} ($\cV_{n}$) and \emph{stereotypical} ($\cV_{s}$),
our proposed method learns a projection that satisfies the following four criteria:
\begin{enumerate}[(i)]
\itemsep=0pt
\item  for $w_f \in \cV_f$, we protect its feminine properties,
\item for $w_m \in \cV_m$, we protect its masculine properties,
\item for $w_n \in \cV_n$, we protect its gender neutrality, and
\item for $w_s \in \cV_s$, we remove its gender biases.
\end{enumerate}

By definition the four word categories are mutually exclusive and the total vocabulary is expressed by their disjunction $\cV = \cV_{m} \cup \cV_{f} \cup \cV_{n} \cup \cV_{s}$.
A key feature of the proposed method that distinguishes it from prior work on debiasing word embeddings is its ability to differentiate between undesirable (stereotypical) biases from the desirable (expected) gender information in words. 
The procedure we follow to compile the four word-sets is described later in \autoref{sec:exp-details}, and the words that belong to each of the four categories are shown in the supplementary material.

To explain the proposed gender debiasing method, let us first consider a \emph{feminine} regressor  $C_f: \R^{l} \rightarrow  [0,1]$, parameterised by $\vec{\theta_f}$, that predicts  the degree of feminineness of the word $w$. 
Here, highly feminine words are assigned values close to 1.
Likewise, let us consider a \emph{masculine} regressor $C_m: \R^{l} \rightarrow  [0,1]$, parametrised by $\vec{\theta_m}$, that predicts the degree of masculinity of $w$.
We then learn the debiasing function as the encoder $E: \R^{d} \rightarrow \R^{l}$ of an autoencoder (parametrised by $\vec{\theta}_{e}$), where the corresponding decoder (parametrised by $\vec{\theta}_{d}$) is given by $D: \R^{l} \rightarrow \R^{d}$. 

For feminine and masculine words, we require the encoded space to retain the gender-related information.
The squared losses, $L_{f}$ and $L_{m}$, given respectively by \eqref{eq:female-loss} and \eqref{eq:male-loss}, express the extent to which this constraint is satisfied.
\par\nobreak
{\small
\begin{align}
\label{eq:female-loss}
L_{f} &= \sum_{w \in \cV_{f}}\norm{C_{f}(E(\vec{w})) - 1}_{2}^{2} + \sum_{w \in \cV\setminus\cV_{f}} \norm{C_{f}(E(\vec{w}))}_{2}^{2} \\
\label{eq:male-loss}
L_{m} &= \sum_{w \in \cV_{m}}\norm{C_{m}(E(\vec{w})) - 1}_{2}^{2} + \sum_{w \in \cV\setminus\cV_{m}} \norm{C_{f}(E(\vec{w}))}_{2}^{2}
\end{align}
}% 
Here, for notational simplicity, we drop the dependence on parameters.

For the stereotypical and gender-neutral words, we require that they are embedded into a subspace that is orthogonal to a gender directional vector, $\vec{v}_{g}$, computed using a set, $\Omega$, of feminine and masculine word-pairs $(w_{f}, w_{m}) (\in \Omega)$ as given by \eqref{eq:gender}.
\begin{align}
\label{eq:gender}
\vec{v}_{g} = \frac{1}{|\Omega|}\sum_{(w_{f}, w_{m}) \in \Omega} \left( E(\vec{w}_{m}) - E(\vec{w}_{f}) \right)
\end{align}
Prior work on gender debiasing~\cite{Tolga:NIPS:2016,Zhao:2018ab} showed that the vector difference between the embeddings for male-female word-pairs such as \emph{he} and \emph{she} accurately represents the gender direction.
When training, we keep $\vec{v}_{g}$ fixed during an epoch, and re-estimate $\vec{v}_{\rm g}$ between every epoch.
We consider the squared inner-product between $\vec{v}_{g}$ and the debiased stereotypical or gender-neutral words as the loss, $L_{g}$, as given by \eqref{eq:gender-loss}.
\begin{align}
\label{eq:gender-loss}
L_{g} = \sum_{w \in \cV_{n} \cup \cV_{s}} {(\vec{v}_{g}\T\vec{w})}^{2}
\end{align}

It is important that we preserve the semantic information encoded in the word embeddings as much as possible  when we perform debiasing.
If too much information is removed from the word embeddings, not limited to gender-biases, then the debiased word embeddings might not be sufficiently accurate to be used in downstream NLP applications.
For this purpose, we minimise the reconstruction loss, $L_{r}$, for the autoencoder given by \eqref{eq:reconst-loss}.
\begin{align}
\label{eq:reconst-loss}
L_{r} = \sum_{w \in \cV} \norm{D(E(\vec{w})) - \vec{w}}_{2}^{2}
\end{align}

Finally, we define the total objective as the linearly-weighted sum of the above-defined losses as given by \eqref{eq:total-loss}. 
\begin{align}
\label{eq:total-loss}
L = \lambda_{f} L_{f} + \lambda_{m} L_{m} + \lambda_{g} L_{g} + \lambda_{r} L_{r} 
\end{align}
Here, the coefficients $\lambda_{f}, \lambda_{m}, \lambda_{g}, \lambda_{r}$ are nonnegative hyper-parameters that add to 1.
They determine the relative importance of the different constraints we consider and can be learnt using training data or determined via cross-validation over a dedicated validation dataset. In our experiments, we use the latter approach.

\subsection{Implementation and Training}
$C_{f}$ and $C_{m}$ are both implemented as feed forward neural networks with one hidden layer and the sigmoid function is used as the nonlinear activation. Increasing the number of hidden layers beyond one for $C_{f}$ and $C_{m}$ did not result in a significant increase in accuracy.
Both the encoder $E$ and the decoder $D$ of the autoencoder are implemented as feed forward neural networks with two hidden layers.
Hyperbolic tangent is used as the activation function throughout the autoencoder.

The objective \eqref{eq:total-loss} is minimised w.r.t. the parameters $\vec{\theta}_{f}$, $\vec{\theta}_{m}$, $\vec{\theta}_{e}$ and $\vec{\theta}_{d}$ for a given pre-trained set of word embeddings.
During optimisation, we used dropout with probability $0.01$ and use stochastic gradient descent with initial learning rate set to $0.1$.
The hyper-parameters $\lambda_{f}, \lambda_{m}, \lambda_{g}, \lambda_{r}$ are estimated using a separate validation dataset as described later in \autoref{sec:exp-details}.

Note that it is possible to pre-train $C_{f}$ and $C_{m}$ separately using $\cV_{f}$ and $\cV_{m}$ prior to training the full objective \eqref{eq:total-loss}. 
In our preliminary experiments, we found that initialising $\vec{\theta}_{f}$ and $\vec{\theta}_{m}$ to the pre-trained versions of $C_{f}$ and $C_{m}$ to be helpful for the optimisation process, resulting in early convergence to better solutions compared to starting from random initialisations for $\vec{\theta}_{f}$ and $\vec{\theta}_{m}$.
For pre-training $C_{f}$ and $C_{m}$ we used Adam optimiser~\cite{Kingma:ICLR:2015} with initial learning rate set to 0.0002 and a mini-batch size of 512. Autoencoder is also pre-trained using a randomly selected 5000 word embeddings and dropout regularisation is applied with probability 0.05.

We note that $\cV_{f}$ and $\cV_{m}$ are separate word sets, not necessarily having corresponding feminine-masculine pairs as in $\Omega$ used in \eqref{eq:gender-loss}. It is of course possible to re-use the words in $\Omega$ in $\cV_{f}$ and $\cV_{m}$, and we follow this approach in our experiments, which helps to decrease the number of seed words required to train the proposed method.
Moreover, the number of training examples across the four categories $\cV_{f}, \cV_{m}, \cV_{n}, \cV_{s}$ were significantly different, which resulted in an imbalanced learning setting. 
We conduct one-sided undersampling~\cite{Kubat:ICML:97} to successfully overcome this data imbalance issue.
The code and the debiased embeddings are publicly available\footnote{\url{https://github.com/kanekomasahiro/gp_debias}}.

\section{Experiments}

%: TODO 
% Tell that even if we do not outperform GN-GloVe, it is not a method that can be applied to pre-trained word embeddings. It learns the debiased embeddings from scratch. Therefore, it has more flexibility compared to GP.
% Swap semantic similarity and analogy

\subsection{Training and Development Data}
\label{sec:exp-details}

We use the feminine and masculine word lists (223 words each) created by~\newcite{Zhao:2018ab} as $\cV_{f}$ and $\cV_{m}$, respectively.
To create a gender-neutral word list, $\cV_{n}$, we select gender-neutral words from a list of 3000 most frequent words in English\footnote{\url{https://bit.ly/2SvBINY}}.
Two annotators independently selected words and subsequently verified for gender neutrality.
The final set of $\cV$ contains 1031 gender-neutral words.
We use the stereotypical word list compiled by \newcite{Tolga:NIPS:2016} as $\cV_{s}$, which contains 166 professions that are stereotypically associated with one type of a gender.
The four sets of words used in the experiments are shown in the supplementary material. 

We train GloVe~\cite{Glove} on 2017 January dump of English Wikipedia to obtain pre-trained $300$-dimensional word embeddings for 322636 unique words. In our experiments, we set both $d$ and $l$ to $300$ to create $300$-dimensional de-biased word embeddings.
We randomly selected 20 words from each of the 4 sets $\cV_{f}$, $\cV_{m}$, $\cV_{n}$ and $\cV_{s}$, and used them as a development set for pre-training $C_{f}$ and $C_{m}$ and to estimate the hyperparameters in \eqref{eq:total-loss}.
The optimal hyperparameter values estimated on this development dataset are: 
$\lambda_{f} = \lambda_{m} = \lambda_{g} = 0.0001$, and $\lambda_{r} = 1.0$.
In our preliminary experiments we observed that increasing $\lambda_{f}, \lambda_{m}$ and $\lambda_{g}$ relative to $\lambda_{r}$ results in higher reconstruction losses in the autoencoder.
This shows that the ability to accurately reconstruct the original word embeddings is an important requirement during debiasing.
%Because the losses are not scaled to a uniform range, the relative magnitudes of these coefficients do not necessarily indicate the importance of the individual components to the overall training objective.

\subsection{Baselines and Comparisons}
We compare our proposed method against several baselines.
\paragraph{GloVe:}  is the pre-trained GloVe embeddings described in \autoref{sec:exp-details}. This baseline denotes a non-debiased version of the word embeddings.
\paragraph{Hard-GloVe:} We use the implementation\footnote{\url{https://github.com/tolga-b/debiaswe}} of hard-debiasing~\cite{Tolga:NIPS:2016} method by the original authors  and produce a debiased version of the pre-trained GloVe embeddings.\footnote{\newcite{Tolga:NIPS:2016} released debiased embeddings for word2vec only and for comparison purposes with GN-GloVe, we use GloVe as the pre-trained word embedding and apply hard-debiasing on GloVe} 
\paragraph{GN-GloVe}: We use debiased GN-GloVe embeddings released by the original authors\footnote{\url{https://github.com/uclanlp/gn_glove}}, without retraining ourselves as a baseline. 
\paragraph{AE (GloVe):} We train an autoencoder by minimising the reconstruction loss defined in \eqref{eq:reconst-loss} and encode the pre-trained GloVe embeddings to a vector space with the same dimensionality. This baseline can be seen as surrogated version of the proposed method with $\lambda_{f} = \lambda_{m} = \lambda_{g} = 0$. \textbf{AE (GloVe)} does \emph{not} perform debiasing and shows the amount of semantic information that can be preserved by autoencoding the input embeddings. 
\paragraph{AE (GN-GloVe):} Similar to \textbf{AE (GloVe)}, this method autoencodes the debiased word embeddings produced by \textbf{GN-GloVe}. 
\paragraph{GP (GloVe):} We apply the proposed \emph{gender-preserving} (\textbf{GP}) debiasing method on pre-trained GloVe embeddings to debias it. 
\paragraph{GP (GN-GloVe):} To test whether we can use the proposed method  to further debias word embeddings that are already debiased using other methods, we apply it on GN-GloVe.

\subsection{Evaluating Debiasing Performance} 
\label{sec:exp:debias}

\begin{table*}[t]
%\small
 \centering
 \begin{tabular}{l ccc ccc}
 \toprule
 \multirow{3}{*}{Embeddings} & \multicolumn{3}{c}{\textsf{SemBias}} & \multicolumn{3}{c}{\textsf{SemBias-subset}} \\
  \cmidrule(lr){2-4}  \cmidrule(lr){5-7} \\
 & Definition $\uparrow$ & Stereotype $\downarrow$ & None $\downarrow$ & Definition $\uparrow$  & Stereotype $\downarrow$ & None $\downarrow$ \\
 \midrule
 GloVe & 80.2 & 10.9 & 8.9 & 57.5 & 20 & 22.5 \\
 Hard-Glove & 84.1 & 9.5 & 6.4 & 25 & 47.5 & 27.5 \\
 GN-GloVe & 97.7 & 1.4 & 0.9 & 75 & 15 & 10 \\
 \bottomrule
 AE (GloVe) & 82.7 & 8.2 & 9.1 & $62.5^{\dagger}$ & $17.5^{\dagger}$ & 20 \\
 AE (GN-GloVe) & $98.0^{\dagger \ast}$ & $1.6^{\dagger \ast}$ & $\mathbf{0.5}^{\dagger \ast}$ & 77.5 & $17.5^{\dagger}$ & $\mathbf{5}^{\dagger \ast}$ \\
 \bottomrule
 GP (GloVe) & $84.3^{\ast}$ & 8.0 & $7.7^{\ast}$ & $65^{\dagger}$ & $15^{\dagger}$ & 20 \\
 %Our GloVe & 85.9 & 6.6 & 7.5 & 67.5 & 12.5 & 20 \\
 GP (GN-GloVe) & $\mathbf{98.4}^{\dagger \ast}$ & $\mathbf{1.1}^{\dagger \ast}$ & $\mathbf{0.5}^{\dagger \ast}$ & $\mathbf{82.5}^{\dagger \ast}$ & $\mathbf{12.5}^{\dagger \ast}$ & $\mathbf{5}^{\dagger \ast}$ \\
 %Our GN-GloVe & \bf 98.2 & \bf 1.1 & \bf 0.7 & \bf 82.5 & \bf 10 & \bf 7.5 \\
 \bottomrule
 \end{tabular}
  \caption{Prediction accuracies for gender relational analogies.  $\ast$ and $\dagger$ indicate statistically significant differences against respectively \textbf{GloVe} and \textbf{Hard-GloVe}.}
  \label{tbl:sembias}
 % \vspace{-5mm}
\end{table*}

We use the \textsf{SemBias} dataset created by \newcite{Zhao:2018ab} to evaluate the level of gender bias in word embeddings.
Each instance in \textsf{SemBias} consists of four word pairs: a gender-definition word pair (\textbf{Definition}; e.g. ``waiter - waitress''),
a gender-stereotype word pair (\textbf{Stereotype}; e.g., ``doctor - nurse'') and two other word-pairs that have similar meanings but not a gender relation (\textbf{None}; e.g., ``dog - cat'', ``cup - lid'').
\textsf{SemBias} contains 20 gender-stereotype word pairs and 22 gender-definitional word pairs and use their Cartesian product to generate 440 instances.
Among the 22 gender-definitional word pairs, 2 word-pairs are not used as the seeds for training.
Following, \newcite{Zhao:2018ab}, to test the generalisability of a debiasing method, we use the subset (\textsf {SemBias-subset}) of 40 instances associated with these 2 pairs.
We measure relational similarity between $(he, she)$ word-pair and a word-pair $(a, b)$ in \textsf{SemBias} using the cosine similarity between the $\vv{he} - \vv{she}$ gender directional vector and $\vec{a} - \vec{b}$ using the word embeddings under evaluation.
For the four word-pairs in each instance in \textsf{SemBias}, we select the word-pair with the highest cosine similarity with $\vv{he} - \vv{she}$ as the predicted answer. 
In \autoref{tbl:sembias}, we show the percentages where a word-pair is correctly classified as \textbf{Definition}, \textbf{Stereotype}, or \textbf{None}. 
If the word embeddings are correctly debiased, we would expect a high accuracy for \textbf{Definitions} and low accuracies for \textbf{Stereotypes} and \textbf{Nones}.

% Glove contains biases. hard glove does not remove much biases but GN-GloVe can do so.
% GP can reduce bias in GloVe better than hard debiasing (hard-glove)
% By applying GP on the debiased embeddings learnt by GN-GloVe we obtain best performance on both SemBias and SemBias (subset).
% This is nice because it shows that it can further debias GN-GloVe
% GN-Glove overfits but this issue is resolved by applying GP.

From \autoref{tbl:sembias}, we see that the best performances (highest accuracy on \textbf{Definition} and lowest accuracy on \textbf{Stereotype}) are reported by \textbf{GP (GN-GloVe)}, which is the application of the proposed method to debias word embeddings learnt by \textbf{GN-GloVe}.
In particular, in both \textsf{SemBias} and \textsf{SemBias-subset}, \textbf{GP (GN-GloVe)} statistically significantly outperforms \textbf{GloVe} and \textbf{Hard-Glove} according to Clopper-Pearson confidence intervals~\cite{Clopper:1934}.
Although \textbf{GN-GloVe} obtains high performance on \textsf{SemBias}, it does not generalise well to \textsf{SemBias-subset}.
However, by applying the proposed method, we can further remove any residual gender biases from \textbf{GN-GloVe}, which shows that the proposed method can be applied in conjunction with \textbf{GN-GloVe}.
We see that \textbf{GloVe} contains a high percentage of stereotypical gender biases, which justifies the need for debiasing methods.
By applying the proposed method on \textbf{GloVe} (corresponds to \textbf{GP (GloVe)}) we can decrease the gender biases in \textbf{GloVe}, while preserving useful gender-related information for detecting definitional word-pairs.
Comparing corresponding \textbf{AE} and \textbf{GP} versions of \textbf{GloVe} and \textbf{GN-GloVe}, we see that autoencoding alone is insufficient to consistently preserve gender-related information.

\subsection{Preservation of Word Semantics}
\label{sec:exp:semantics}

\begin{table*}[t]
%\small
\centering
\begin{tabular}{l rrrrr}
\toprule
Embeddings & sem & syn & total & MSR & SE \\
\midrule
GloVe & 80.1 & 62.1 & 70.3 & 53.8 & 38.8 \\
Hard-GloVe & 80.3 & \bf 62.7 & \bf 70.7 & \bf 54.4 & 39.1 \\
GN-GloVe & 77.8 & 60.9 & 68.6 & 51.5 & 39.1 \\
\midrule
AE (GloVe) & \bf 81.0 & 61.9 & 70.5 & 52.6 & 38.9 \\
AE (GN-GloVe) & 78.6 & 61.3 & 69.2 & 51.2 & 39.1 \\
\midrule
GP (GloVe) & 80.5 & 61.0 & 69.9 & 51.3 & 38.5 \\
GP (GN-GloVe) & 78.3 & 61.3 & 69.0 & 51.0 & \bf 39.6\\
\bottomrule
\end{tabular}
\caption{Accuracy for solving word analogies.}
\label{tbl:analogy}
\end{table*}

It is important that the debiasing process removes only gender biases and preserve other information unrelated to gender biases in the original word embeddings. If too much information is removed from word embeddings during the debiasing process, then the debiased embeddings might not carry adequate information for downstream NLP tasks that use those debiased word embeddings.

To evaluate the semantic accuracy of the debiased word embeddings, following prior work on debiasing~\cite{Tolga:NIPS:2016,Zhao:2018aa},
 we use them in two popular tasks: semantic similarity measurement and analogy detection. 
We recall that we do \emph{not} propose novel word embedding learning methods in this paper, and what is important here is whether the debiasing process preserves as much information as possible in the original word embeddings.

\subsubsection{Analogy Detection}
\label{sec:analogy}

Given three words $a, b, c$ in analogy detection, we must predict a word $d$ that completes the analogy ``$a$ is $b$ as $c$ is to $d$''.
We use the CosAdd~\cite{Levy:CoNLL:2014} that finds $\vec{d}$ that has the maximum cosine similarity with $(\vec{b} - \vec{a} + \vec{c})$.
We use the semantic (\textbf{sem}) and syntactic (\textbf{syn}) analogies in the Google analogy dataset~\cite{NIPS2013_5021} (in \textbf{total} contains 19,556 questions), \textbf{MSR} dataset (7,999 syntactic questions)~\cite{mikolov-yih-zweig:2013:NAACL-HLT} and SemEval dataset (\textbf{SE}, 79 paradigms)~\cite{S12-1047} as benchmark datasets.
The percentage of correctly solved analogy questions is reported in \autoref{tbl:analogy}.
We see that there is no significant degradation of performance due to debiasing using the proposed method.

\subsubsection{Semantic Similarity Measurement}
\label{sec:similarity}

\begin{table}[t]
%\small
 \centering
 \begin{tabular}{l c c}
 \toprule
 Datasets & \#Orig & \#Bal \\
 \midrule
 WS & 353 & 366 \\
 RG & 65 & 77 \\
 MTurk & 771 & 784 \\
 RW & 2,034 & 2,042 \\
 MEN & 3,000 & 3,122 \\
 SimLex & 999 & 1,043 \\
 \bottomrule
 \end{tabular}
  \caption{Number of word-pairs in the original (\textbf{Orig}) and balanced (\textbf{Bal}) similarity benchmarks.}
  \label{tbl:balanced}
\end{table}

\begin{table*}[t!]
%\small
\centering
\begin{tabular}{l rr rr rr rr rr rr}
\toprule
\multirow{2}{*}{Embeddings} & \multicolumn{2}{c}{WS} & \multicolumn{2}{c}{RG} & \multicolumn{2}{c}{MTurk} & \multicolumn{2}{c}{RW} & \multicolumn{2}{c}{MEN} & \multicolumn{2}{c}{SimLex} \\
& Orig & Bal & Orig & Bal & Orig & Bal & Orig & Bal & Orig & Bal & Orig & Bal  \\
\midrule
GloVe           & 61.6 & 62.9       & 75.3 & 75.5       & 64.9 & 63.9       & 37.3 & 37.5       & 73.0 & 72.6       & 34.7 & 35.9 \\
Hard-GloVe      & 61.7 & 63.1       & 76.4 & 76.7       & 65.1 & 64.1       & 37.4 & 37.4       & 72.8 & 72.5       & 35.0 & 36.1 \\
GN-GloVe        & 62.5 & 63.7 & 74.1 & 73.7       & 66.2 & 65.5       & 40.0 & 40.1   & 74.9 & 74.5 & 37.0 & 38.1 \\
\midrule
AE (GloVe)      & 61.3 & 62.6       &\bf 77.1 &\bf 76.8       & 64.9 & 64.1       & 35.7 & 35.8       & 71.9 & 71.5       & 34.7 & 35.9 \\
AE (GN-GloVe)   & 61.3 & 62.6       & 73.0 & 74.0       & 66.3 & 65.5   & 38.7 & 38.9       & 73.8 & 73.4       & 36.7 & 37.7 \\
\midrule
GP (GloVe)      & 59.7 & 61.0	    & 75.4 & 75.5	    & 63.9 & 63.1       & 34.7 & 34.8     & 70.8 & 70.4       & 33.9 & 35.0  \\
GP (GN-GloVe)   & \bf 63.2 & \bf 64.3 	& 72.2 &  72.2  & \bf 67.9 & \bf 67.4 	& \bf 43.2 &  \bf 43.3	& \bf 75.9 & \bf 75.5 	& \bf 38.4 & \bf 39.5 \\
\bottomrule
\end{tabular}
\caption{Spearman correlation between human ratings and cosine similarity scores computed using word embeddings for the word-pairs in the original and balanced versions of the benchmark datasets.}
\label{tbl:sim}
\end{table*}

\begin{figure*}[t!]
\centering
	\begin{subfigure}[b]{0.49\textwidth}
		\centering
		\includegraphics[height=2.5in]{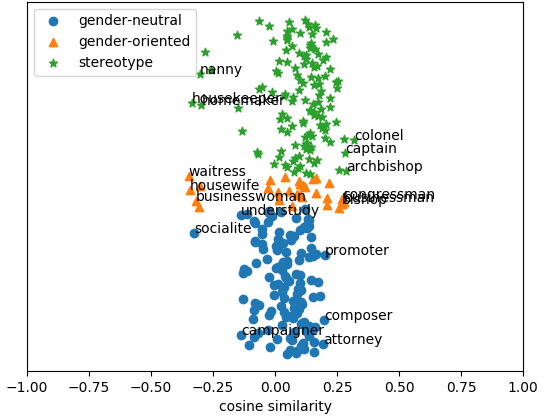}
		\caption{GloVe}
		\label{fig:glove}
	\end{subfigure}
	\begin{subfigure}[b]{0.49\textwidth}
		\centering
		\includegraphics[height=2.5in]{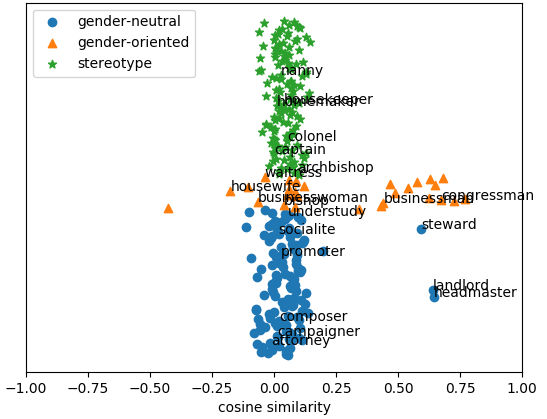}
		\caption{GN (GloVe)}
	\end{subfigure}
	\vskip\baselineskip
	\begin{subfigure}[b]{0.49\textwidth}
		\centering
		\includegraphics[height=2.5in]{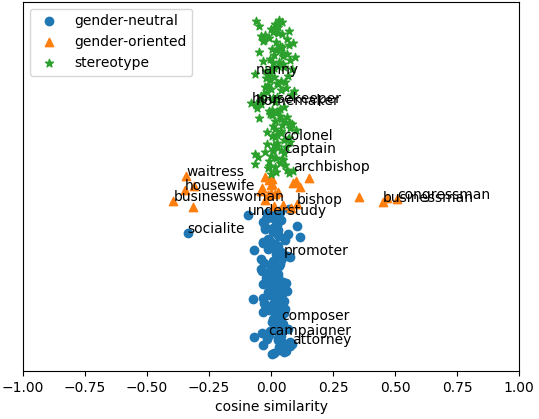}
		\caption{Hard-Glove}
	\end{subfigure}
	\begin{subfigure}[b]{0.49\textwidth}
		\centering
		\includegraphics[height=2.5in]{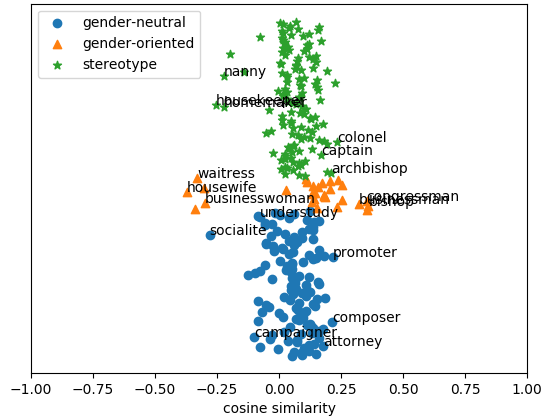}
		\caption{GP (GloVe)}
	\end{subfigure}
\caption{Cosine similarity between gender, gender-neutral, stereotypical words and the gender direction.}
\label{fig:plots}
%\vspace{-5mm}
\end{figure*}

The correlation between the human ratings and similarity scores computed using word embeddings for pairs of words has been used as a measure of the quality of the word embeddings~\cite{mikolov-yih-zweig:2013:NAACL-HLT}.
We compute cosine similarity between word embeddings and measure Spearman correlation against human ratings for the word-pairs in the following benchmark datasets: Word Similarity 353 dataset ({\bf WS})~\cite{Finkelstein:2001}, Rubenstein-Goodenough dataset ({\bf RG})~\cite{Rubenstein:1965}, {\bf MTurk}~\cite{Halawi:2012}, rare words dataset ({\bf RW})~\cite{W13-3512}, {\bf MEN} dataset~\cite{P12-1015} and {\bf SimLex} dataset~\cite{J15-4004}.

Unfortunately, existing benchmark datasets for semantic similarity were not created considering gender-biases and contain many stereotypical examples. 
For example, in \textbf{MEN}, the word \emph{sexy} has high human similarity ratings with \emph{lady}  and \emph{girl} compared to \emph{man} and \emph{guy}.
Furthermore, masculine words and \emph{soldier} are included in multiple datasets with high human similarity ratings, whereas it is not compared with feminine words in any of the datasets.
Although prior work studying gender bias have used these datasets for evaluation purposes~\cite{Tolga:NIPS:2016,Zhao:2018aa}, we note that high correlation with human ratings can be achieved with biased word embeddings.

To address this issue, we \emph{balance} the original datasets with respect to gender by including extra word pairs generated from the opposite sex with the same human ratings.
For instance, if the word-pair (\emph{baby}, \emph{mother}) exists in the dataset, we add a new pair (\emph{baby}, \emph{father}) to the dataset. Ideally, we should re-annotate this balanced version of the dataset to obtain human similarity ratings.
However, such a re-annotation exercise would be costly and inconsistent with the original ratings.
Therefore, we resort to a proxy where we reassign the human rating for the original word-pair to its derived opposite gender version.
\autoref{tbl:balanced} shows the number of word-pairs in the original (\textbf{Orig}) and balanced (\textbf{Bal}) similarity benchmarks.

As shown in \autoref{tbl:sim}, \textbf{GP (GloVe)} and \textbf{GP (GN-GloVe)} obtain the best performance on the balanced versions of all benchmark datasets. 
Moreover, the performance of \textbf{GP (GloVe)} on both original and balanced datasets is comparable to that of \textbf{GloVe}, which indicates that the information encoded in GloVe embeddings are preserved in the debiased embeddings, while removing stereotypical gender biases. The autoencoded versions report similar performance to the original input embeddings.

Overall, the results on the analogy detection and semantic similarity measurement tasks show that our proposed method removes only gender-biases and preserve other useful gender-related information.

\subsection{Visualising the Effect of Debiasing}
\label{sec:visual}

To visualise the effect of debiasing on different word categories, we compute the cosine similarity between the gender directional vector $\vv{he}-\vv{she}$, and selected \emph{gender-oriented} (female or male), \emph{gender-neutral} and stereotypical words.
In \autoref{fig:plots}, horizontal axises show the cosine similarity with the gender directional vector (positive scores for masculine words) and the words are alphabetically sorted within each category.

From \autoref{fig:plots}, we see that the original \textbf{GloVe} embeddings show a similar spread of cosine similarity scores for gender-oriented as well as stereotypical words. When debiased by hard-debias (\textbf{Hard-GloVe}) and \textbf{GN-GloVe}, we see that stereotypical and gender-neutral words get their gender similarity scores equally reduced. 
Interestingly, \textbf{Hard-GloVe} shifts even gender-oriented words towards the masculine direction.
On the other hand, \textbf{GP (GloVe)} decreases gender bias in the stereotypical words, while almost preserving gender-neutral and gender-oriented words as in \textbf{GloVe}.

Considering that a significant number of words in English are gender-neutral, it is essential that debiasing methods do not adversely change their orientation.
In particular, the proposed method's ability to debias stereotypical words that carry unfair gender-biases, while preserving the gender-orientation in feminine, masculine and neutral words is important when applying the debiased word embeddings in NLP applications that depend on word embeddings for representing the input texts

\section{Conclusion}
We proposed a method to remove gender-specific biases from pre-trained word embeddings. 
Experimental results on multiple benchmark datasets demonstrate that the proposed method can accurately debias pre-trained word embeddings, outperforming previously proposed debiasing methods, while preserving useful semantic information.
In future, we plan to extend the proposed method to debias other types of demographic biases such as ethnic, age or religious biases. 
%A straightforward extension of the proposed method would be to create predictors for each value of the protected attribute and then require stereotypical instances to be orthogonal to the space defined by the directions of those values.
%We defer such extensions to future work.
%Debiasing word embeddings that are used as features for representing inputs in NLP systems is an important first step towards making NLP systems unbiased. However, even if we use debiasing word embeddings, NLP systems can still become biased if the training data contain gender or other types of demographic biases. We hope our work will inspire the NLP community to study this important problem further.

\bibliography{UnbiasedACL}
\bibliographystyle{acl_natbib}

\end{document}